\documentclass[11pt,a4paper]{article}
\usepackage[colorlinks=true,linkcolor=blue]{hyperref}

\usepackage{cmap}
\usepackage[utf8x]{inputenc}
\usepackage[T1]{fontenc}
\usepackage[english]{babel}
\usepackage{amssymb,amsmath}
\usepackage{amsthm}
\usepackage{bbm}
\usepackage{microtype}
\usepackage{lmodern}

{
\rmfamily
\DeclareFontShape{T1}{lmr}{m}{sc}{<->ssub*cmr/m/sc}{}
\DeclareFontShape{T1}{lmr}{b}{sc}{<->ssub*cmr/b/sc}{}
\DeclareFontShape{T1}{lmr}{bx}{sc}{<->ssub*cmr/bx/sc}{}
}

\newcommand{\thmheadercommand}[1]{\textbf{\scshape{}#1.\\*}}

\newtheoremstyle{yannthm}{\topsep}{\topsep}{\slshape}{}{\scshape\bfseries}{.}{.5em}{%
\thmname{#1}\thmnumber{ #2}\thmnote{#3}%
}
\newtheoremstyle{yannthm2}{\topsep}{\topsep}{}{}{\scshape\bfseries}{.}{.5em}{%
\thmname{#1}\thmnumber{ #2}\thmnote{#3}%
}

\def\d{\operatorname{d}\!{}}

\def\R{{\mathbb{R}}}

\renewcommand{\leq}{\leqslant}

\newcommand{\deq}{\mathrel{\mathop:}=}

\def\eps{\varepsilon}
\renewcommand{\epsilon}{\varepsilon}
\renewcommand{\phi}{\varphi}

\DeclareMathOperator{\Var}{Var}

\let\oldPr\Pr
\renewcommand{\Pr}{\oldPr\nolimits}
\newcommand{\E}{\mathbb{E}}

\DeclareMathOperator{\Tr}{Tr}

\DeclareMathOperator{\Id}{Id}

\newcommand{\norm}[1]{\left\|#1\right\|}
\newcommand{\scal}[2]{\left< \, #1 \mid #2 \, \right>}

{
\theoremstyle{yannthm}
\newtheorem{defi}{Definition}
\newtheorem*{defi*}{Definition}
\newtheorem{prop}[defi]{Proposition}
\newtheorem*{prop*}{Proposition}

\newtheorem*{thm*}{Theorem}

\newtheorem*{lem*}{Lemma}

\newtheorem*{cor*}{Corollary}

\newtheorem*{ex*}{Example}

\newtheorem*{subenonce}{}

\theoremstyle{yannthm2}

\newtheorem*{exo*}{Exercise}

\newtheorem*{rem*}{Remark}

\newtheorem*{subenonce2}{}

}

\newcommand{\transp}[1]{#1^{\!\top}\!}

\usepackage{algorithm2e}
\usepackage{graphicx}
\usepackage{xcolor}

\newcommand{\authorcomment}[2]{{\color[rgb]{#1}#2}}

\newcommand{\NDY}[1]{\authorcomment{0.0,0.8,0.4}{[NdY: #1]}}
\newcommand{\new}[1]{\authorcomment{0.0,0.0,0.8}{#1}} %for recent changes to the text

\renewcommand{\authorcomment}[2]{}  %delete comments in final version
\renewcommand{\new}[1]{#1}  %remove coloring for final version

\DeclareMathOperator{\Diag}{Diag}

\title{Training recurrent networks online\\without backtracking}

\author{Yann Ollivier \and Corentin Tallec \and Guillaume Charpiat}

\date{}

\begin{document}

\maketitle

\begin{abstract}
We introduce the ``NoBackTrack'' algorithm to train the parameters of
dynamical systems such as recurrent neural networks. This algorithm works
in an online, memoryless setting, thus requiring no backpropagation
through time, and is scalable, avoiding the large computational and memory cost of
maintaining the full gradient of the current state with respect to the
parameters.

The algorithm essentially maintains, at each time, a single search
direction in parameter space. The evolution of this search direction is
partly stochastic and is constructed in such a way to provide, at every time, an
\emph{unbiased} random estimate of the gradient of the loss function with
respect to the parameters. Because the gradient estimate is unbiased,
on average over time the parameter is updated as it should.

The resulting gradient estimate can then be fed to a lightweight
Kalman-like filter to yield an improved algorithm.
For recurrent neural networks, the resulting algorithms scale linearly
with the number of parameters.

\new{Small-scale experiments confirm the suitability of the approach,
showing that the stochastic approximation
of the gradient introduced in the algorithm is not detrimental to
learning.
In particular, the Kalman-like version of
NoBackTrack is superior to backpropagation through time (BPTT) when the
time span of dependencies in the data is longer than the truncation span
for BPTT.}
% Preliminary tests on a simple task show that the stochastic approximation
% of the gradient introduced in the algorithm does not seem to introduce
% too much noise in the trajectory, compared to maintaining the full
% gradient, and confirm the good performance and scalability of the Kalman-like version of
% NoBackTrack.
\end{abstract}

Consider the problem of training the parameters $\theta$ of a dynamical
system over a variable $h\in \R^n$ subjected to the evolution equation
\begin{equation}
\label{eq:evol}
h(t+1)=f(h(t),x(t),\theta)
\end{equation}
where $f$ is a fixed function of $h$ and of an input signal 
$x(t)$, depending on parameters $\theta$. The goal is online minimization of a loss
function
$\sum_t \ell_t(\hat y(t),y(t))$ between a desired output $y(t)$ at time $t$
and a prediction\footnote{The prediction $\hat{y}$ may not live in the
same set as $y$. Often, $\hat y$ encodes a probability
distribution over the possible values of $y$, and the loss is the logarithmic
loss $\ell=-\log p_{\hat y}(y)$.}
\begin{equation}
\hat y(t)=Y(h(t),\phi)
\end{equation}
computed from $h(t)$ and additional parameters $\phi$. \NDY{One could
subsume $\phi$ inside $\theta$ by also subsuming $\hat y$ inside $h$
(ie for an NN, one would consider the non-recurrent softmax output layer as
part of the NN) but that would result in a different algorithm since
$\partial \hat y/\partial h$ would become rank-one-approximated too.}

A typical example we have in mind is a recurrent neural
network, with activities $a_i(t)\deq \mathrm{sigm}(h_i(t))$ and evolution
equation $h_i(t+1)=b_i+\sum_k r_{ki} x_k(t)+\sum_j W_{ji} a_j(t)$, with
parameter $\theta=(b_i,r_{ki},W_{ji})_{i,j,k}$.

If the full target sequence $y(t)_{t\in [0;T]}$ is known in
advance, one strategy is to use the backpropagation through time
algorithm (BPTT, see e.g. \cite{Jaeger_tutorial}) to compute the gradient of the total loss $L_T\deq
\sum_{t=0}^T \ell_t$ with respect to the parameters $\theta$ and $\phi$,
and use gradient descent on $\theta$ and $\phi$.

However, if the data $y(t+1)$ arrive one at a time in a streaming
fashion, backpropagation through time would require making a full
backward computation from time $t+1$ to time $0$ after each new data
point becomes available. This results in an $\Omega(t^2)$ complexity and
in the necessity to store past states, inputs, and outputs. A possible
strategy is to only backtrack by a finite number of time steps
\cite{Jaeger_tutorial} rather than going back all the way to $t=0$. But
this provides biased gradient estimates and may impair detection of time
dependencies with a longer range than the backtracking time range.

By contrast, methods which are fully online are typically not scalable.
One strategy, known as \emph{real-time recurrent learning} (RTRL) in the
recurrent network community,\footnote{This amounts to applying
forward automatic differentiation.} maintains the full gradient of the
current state with respect to the parameters:
\begin{equation}
G(t)\deq \frac{\partial h(t)}{\partial \theta}
\end{equation}
which satisfies the evolution equation
\begin{equation}
\label{eq:Gevol}
G(t+1)=\frac{\partial f(h(t),x(t),\theta)}{\partial
h}\,G(t)+\frac{\partial f(h(t),x(t),\theta)}{\partial \theta}
\end{equation}
(by differentiating \eqref{eq:evol}). Knowing $G(t)$ allows to minimize
the loss via a
stochastic gradient descent on the parameters $\theta$,
 namely,\footnote{We
use the standard convention for Jacobian matrices, namely,
$\partial x/\partial y$ is the matrix with entries $\partial x_i/\partial
y_j$. Then the chain rule writes $\frac{\partial x}{\partial
y}\frac{\partial y}{\partial z}=\frac{\partial x}{\partial z}$. This makes the derivatives $\partial \ell_t/\partial
\theta$ into \emph{row} vectors so that gradient descent is
$\theta\gets\theta-\transp{(\partial \ell_t/\partial \theta)}$.}
\begin{equation}
\label{eq:stochgrad}
\theta\gets \theta-\eta_t \transp{\frac{\partial \ell_t}{\partial
\theta}}
\end{equation}
with learning rate $\eta_t$. Indeed,
the latter quantity can be computed from $G_t$ and from the way the
predictions depend on $h(t)$, via the chain rule
\begin{equation}
\frac{\partial{\ell_t}}{\partial \theta}=\frac{\partial
\ell_t(Y(h(t),\phi),y(t))}{\partial
h}\,G(t)
\end{equation}

However, the full gradient $G(t)$ is an object of dimension $\dim h\times
\dim \theta$. This prevents computing or even storing $G(t)$ for
moderately large-dimensional dynamical systems, such as recurrent neural
networks.

Algorithms using a Kalman filter on $\theta$ also\footnote{One may 
use Kalman filtering either on $\theta$ alone or on the pair $(\theta,h)$. In the first
case, $\frac{\partial \ell_t}{\partial \theta}$ is explicitly needed. In
the second case, all the information about how $\theta$ influences the
current state $h(t)$ is contained in the covariance between $\theta$ and
$h$, which the algorithm must maintain, and which is as costly as
maintaining $G(t)$ above.} rely on this derivative
$\frac{\partial \ell_t}{\partial \theta}$ (see
\cite{Haykin_book,Jaeger_tutorial} for the case of recurrent networks).
So any efficient way of estimating this derivative can be fed, in turn,
to a Kalman-type algorithm.

Algorithms suggested to train hidden Markov models online
(e.g., \cite{Cappe_onlineHMM}, based on expectation-maximization instead
of gradient descent) share the same algebraic structure and suffer from
the same problem.

\section{The NoBackTrack algorithm}

\subsection{The rank-one trick: an expectation-preserving reduction}

We propose to build an approximation $\tilde G(t)$ of $G(t)$ with a more
sustainable algorithmic cost; $\tilde G(t)$ will be random with the
property $\E \tilde G(t)=G(t)$ for all $t$. Then the stochastic
gradient~\eqref{eq:stochgrad} based on $\tilde G(t)$ will introduce
noise, but \emph{no bias}, on the learning of $\theta$: the average
change in $\theta$ after a large
number of time steps will reflect the true gradient direction.
(This is true only if the
noises on $\tilde G(t)$ at different times $t$ are sufficiently
decorrelated. This is the case if the dynamical system \eqref{eq:evol} is
sufficiently ergodic.) Such unbiasedness does not hold, for instance, if
the gradient estimate is simply projected onto the nearest small-rank or
diagonal plus small-rank approximation.\footnote{We tried such methods
first, with less satisfying results. In practice, consecutive projections
%in different directions
tend to interact badly and reduce too much the
older contributions to the gradient.}

The construction of an unbiased $\tilde G$ is based on the following
``rank-one trick''.

\NDY{Is it too much to call this a proposition? What about ``Remark''?}

\begin{prop}[ (Rank-one trick)]
\label{prop:rk1}
Given a
decomposition of a matrix $A$ as a sum of rank-one outer products,
$A=\sum_i v_i \transp{w_i}$, and independent uniform random signs $\eps_i\in
\{-1,1\}$, then
\begin{equation}
\label{eq:Atilde}
\tilde A\deq 
\left(\textstyle{\sum_i} \eps_i
v_i\right)\transp{\left(\textstyle{\sum_j} \eps_j w_j\right)}
\end{equation}
satisfies
\begin{equation}
\E\tilde A=\sum_i v_i\transp{w_i}=A
\end{equation}
that is, $\tilde A$ is an expectation-preserving rank-one approximation
of $A$.

Moreover, one can minimize the variance of $\tilde A$ by taking
advantage of additional degrees of freedom in this decomposition, namely,
one may first replace $v_i$ and $w_i$ with $\rho_i v_i$ and $w_i/\rho_i$
for 
any $\rho_i\in \R^*$. The
choice of $\rho_i$ which yields minimal variance of $\tilde A$ is
when the norms of $v_i$ and $w_i$ become equal, namely,
$\rho_i=\sqrt{\norm{w_i}/\norm{v_i}}$.
\end{prop}

The proof of the first statement is immediate.

The statement about minimizing variance is proven in
Appendix~\ref{app:varproof}.
Minimizing variance thanks to $\rho_i$ is quite important in practice,
see Section~\ref{sec:examples}.

The rank-one trick also extends to tensors of arbitrary order; this may
be useful in more complex situations.\footnote{The most symmetric way to
do this is to use complex roots of unity, for instance, $\sum_i u_i
\otimes v_i \otimes w_i = \E \,\mathrm{Re} \left(
(\sum_i \zeta_i u_i)(\sum_j \zeta_j v_j)(\sum_k \zeta_k w_k)
\right)$ where each $\zeta_i$ is taken independently at random among
$\{1,\mathrm{e}^{\pm 2i\pi/3}\}$. This involves complex numbers but there
is no need to complexify the original dynamical system~\eqref{eq:evol}.
Another, complex-free possibility is to apply the rank-one trick
recursively to tensors of smaller order, for instance, $\sum_i u_i
\otimes v_i \otimes w_i \otimes x_i = \sum_i (u_i\otimes v_i)\otimes
(w_i\otimes x_i)=\E\left[
(\sum_i \eps_i u_i\otimes v_i)(\sum_j \eps_j w_j \otimes x_j)
\right]
$ and then apply \emph{independent} rank-one decompositions in turn to
$\sum_i \eps_i u_i\otimes v_i$ and to $\sum_j \eps_j w_j \otimes x_j$.}

The rank-one reduction $\tilde A$ depends, not only on the value of $A$,
but also on the way $A$ is decomposed as a sum of rank-one terms. In the
applications to recurrent networks below, there is a natural such choice.\footnote{The rank-one
trick may also be performed using random Gaussian vectors, namely $A=\E[
\xi (\transp{\xi}\Sigma^{-1}A)]$ with $\xi=\mathcal{N}(0,\Sigma)$. This
version does not depend on a chosen decomposition of $A$, but depends
on a choice of $\Sigma$. Variance can be much larger in this case: for
instance, if $A=v\transp{w}$ is actually rank-one, then $(\eps
v)(\eps\transp{w})=v\transp{w}$ so that the rank-one trick with random signs is
exact, whereas the Gaussian version yields $(\xi\transp{\xi}\Sigma^{-1}) v\transp{w}$
which is correct only in expectation. This case is particularly relevant
because we are going to apply a reduction at
each time step, thus working on objects that stay close to rank-one. The generalization to tensors is
also more cumbersome in the Gaussian case.}

\bigskip

We use this reduction operation at each step of the dynamical system, to 
build an approximation $\tilde G$ of $G$. A key property is that the evolution equation \eqref{eq:Gevol}
satisfied by $G$ is \emph{affine}, so that if $\tilde G(t)$ is an
unbiased estimate of $G(t)$, then $
\frac{\partial f(h(t),x(t),\theta)}{\partial
h}\,\tilde G(t)+\frac{\partial f(h(t),x(t),\theta)}{\partial \theta}$ is
an unbiased estimate of $G(t+1)$.

This leads to the NoBackTrack algorithm (Euclidean version) described in
Algorithm~\ref{alg:eucl}.
At each step, this algorithm maintains an approximation of $G$ as
\begin{equation}
\label{eq:Gtilde}
\tilde G=\bar
v\transp{\bar w}+\sum_i e_i \transp{w_i}
\end{equation}
where $e_i$ is the $i$-th basis
vector in space $h$, and $w_i\deq \transp{\frac{\partial f_i}{\partial
\theta}}$ are sparse vectors.

To understand this structure, say
that $\tilde G(t-1)=\bar v\transp{\bar w}$ is a rank-one
unbiased approximation of $G(t-1)$. Then the evolution equation
\eqref{eq:Gevol} for $G$ yields $\left(\frac{\partial f}{\partial h}
\right)\left(\bar
v\transp{\bar w}\right) +\frac{\partial f}{\partial
\theta}=\left(\frac{\partial f}{\partial h}
\bar
v\right)\transp{\bar w} +\sum_i e_i \frac{\partial f_i}{\partial \theta}$ as an
approximation of $\tilde G(t)$. This new approximation is not rank-one
any more, but it can be used to perform a gradient step on $\theta$, and
then reduced to a rank-one approximation before the next
time step.

Note that handling $\frac{\partial f_i}{\partial \theta}$ is usually
cheap: in many situations, only a small subset of the parameter
$\theta$ directly influences each component $h_i(t+1)$ given $h(t)$, so
that for each component $i$ of the state space, $\frac{\partial
f_i}{\partial \theta}$ has few non-zero components. For instance, for a
recurrent neural network with activities $a_i(t)\deq \mathrm{sigm}(h_i(t))$ and
evolution equation
$h_i(t+1)=b_i+\sum_k r_{ik} x_k(t)+\sum_j W_{ji} a_j(t)$, the derivative
of $h_i(t+1)$ with respect to the parameter $\theta=(b,r,W)$ only
involves the parameters $b_i$, $r_{ik}$, $W_{ji}$ of unit $i$. In such
situations, the total cost of computing and storing all the $w_i$'s is of the
same order as the cost of computing $h(t+1)$ itself. See
Section~\ref{sec:examples} for details on this example.

\SetKw{Params}{Parameters:}
\SetKw{Maintains}{Maintains:}
\SetKw{Subroutines}{Subroutines:}

\begin{algorithm}
\Params{$h(0)$ (initial state), $\theta_0$, $\phi_0$ (initial value of
the internal and output parameters), $\eta_t$ (learning rate
scheme)\;}
\KwData{$x(t)$ (input signal), $y(t)$ (output signal)\;}
\Maintains{$h(t)$ (current state), $\theta$, $\phi$ (internal and output
parameters), $\bar v$ (column
vector of size $\dim h$), $\bar w$ (column vector of size $\dim \theta$),
$w_i$ (sparse column vectors of size $\dim \theta$) for
$i=1,\ldots,\dim h$.}
\BlankLine

\textbf{Initialization:} $\theta\gets \theta_0$, $\phi\gets \phi_0$,
$\bar v \gets 0$, $\bar w\gets 0$, $w_i\gets 0$\;

\For{$t=0$ \KwTo end-of-time}{

\textbf{Observation step:} Compute prediction $\hat y(t)= Y(h(t),\phi)$ from current state $h(t)$.

Observe $y(t)$ and incur loss
$\ell_t(\hat y(t),y(t))$.

\textbf{Update step:}
Compute derivative of loss with respect to output parameters, $\frac{\partial
\ell_t}{\partial \phi}= \frac{\partial \ell_t(Y(h(t),\phi),y(t))}{\partial
\phi}$, and
update output parameters:
\begin{equation}
\phi\gets \phi-\eta_t \,\transp{\frac{\partial
\ell_t}{\partial \phi}}
\end{equation}

Compute derivative of loss with respect to current state, 
\begin{equation}
H\gets
\frac{\partial{\ell_t\left(Y(h(t),\phi),y(t)\right)}}{\partial h}
\end{equation}

Update internal parameters $\theta$:
\begin{equation}
\theta\gets \theta-\eta_t \,
(Hv) \bar w -\eta_t \,{\textstyle{\sum_i}} H_i w_i
\end{equation}
(this is a gradient step $\theta\gets \theta-\eta_t
\transp{(H\tilde G)}$ using the current gradient estimate $\tilde G$ from
\eqref{eq:Gtilde}).

\textbf{Reduction step:} Draw independent uniform random signs $\eps_i=\pm 1$.
Let $e_i$ be the $i$-th basis vector in state space.
Compute $\bar \rho\deq \sqrt{\norm{\bar w}/\norm{\bar v}}$ and
$\rho_i\deq\sqrt{\norm{w_i}/\norm{e_i}}$ for each $i$. Update
\begin{align}
\bar v & \gets \bar\rho \bar v+{\textstyle \sum_i} \eps_i \rho_i e_i
\\
\bar w & \gets \bar w/\bar\rho+{\textstyle \sum_i} \eps_i w_i/\rho_i
\\
w_i &\gets 0
\end{align}

\textbf{Transition step:}
Observe new value of input signal $x(t)$ and compute next state
$h(t+1)=f(h(t),x(t),\theta)$. Update estimate $\tilde G$:
\begin{align}
\bar v &\gets \frac{\partial f(h(t),x(t),\theta)}{\partial h} \,\bar v
\\
w_i &\gets \transp{\frac{\partial f_i(h(t),x(t),\theta)}{\partial
\theta}}
\\t&\gets t+1
\end{align}

}

\caption{NoBackTrack algorithm, Euclidean version.}
\label{alg:eucl}
\end{algorithm}

\bigskip

After the reduction step of Algorithm~\ref{alg:eucl}, $\bar w$ may be interpreted as a ``search
direction'' in parameter space $\theta$, while $\bar v$ is an estimate of
the effect on the current state $h(t)$ of changing $\theta$ in the
direction $\bar w$. The search direction $\bar w$ evolves stochastically,
but not fully at random, over time, so that on average $\bar
v\transp{\bar w}$ is a fair estimate of the actual influence of the
parameter $\theta$.

\new{Note that in Algorithm~\ref{alg:eucl}, the non-recurrent output parameters
$\phi$ are trained according to their exact gradient. The rank-one trick
is used only for the recurrent part of the system.}

\bigskip

By construction, at each step 
of Algorithm~\ref{alg:eucl}, the quantity $\tilde G_t\deq
\bar v\transp{\bar w}+\sum_i e_i \transp{w_i}$ satisfies
$
\E \tilde G_t=\frac{\partial h(t)}{\partial \theta}
$. However, since the value of $\theta$ changes along the algorithm, we
must be careful about the meaning of this statement. Intuitively, 
this derivative with respect to $\theta$ is taken along the actual
trajectory of parameters $\theta_t$ realized by the algorithm.

\newcommand{\thetatraj}{\boldsymbol\theta}

More formally, let $\thetatraj=(\theta_0,\ldots,\theta_t,\ldots)$ be any
sequence of parameters. Let $f$ be any function depending on this
sequence $\thetatraj$, such as the state of the system at time $t$ (all
functions considered below will depend only on a finite initial segment
of $\thetatraj$). Define $\thetatraj+\eps\deq
(\theta_0+\eps,\ldots,\theta_t+\eps,\ldots)$ and say that 
$f$ has derivative $\frac{\partial f}{\partial
\thetatraj}$ with respect to $\thetatraj$ if
$f(\thetatraj+\eps)=f(\thetatraj)+\eps \frac{\partial f}{\partial
\thetatraj}+O(\eps^2)$ for small $\eps$.

Thanks to this convention, the evolution equation \eqref{eq:Gevol} for
the evolution of $G(t)$ holds for any sequence of parameters
$\thetatraj$, with $G(t)$ defined as $\frac{\partial
h(t)}{\partial\thetatraj}$. The following statement is then easily proved
by induction.

\begin{prop}[ (Unbiased rank-one gradient estimate for dynamical systems)]
At each time step $t$, the quantity
$\tilde G_t\deq
\bar v\transp{\bar w}+\sum_i e_i \transp{w_i}$ from
Algorithm~\ref{alg:eucl} is an unbiased estimate of the gradient of the
state of the system with respect to the parameter:
\begin{equation}
\E \tilde G_t=\frac{\partial h(t)}{\partial \thetatraj}
\end{equation}
where $\thetatraj$ is the sequence of parameters produced by the algorithm.
\end{prop}

\new{
In particular, for learning rates $\eta$ tending to $0$, the parameter
evolves slowly so that the derivative
$\frac{\partial h(t)}{\partial \thetatraj}$ is close to a derivative with
respect to the current value $\theta_t$ of the parameter. Thus, in this
regime, $\frac{\partial h(t)}{\partial \thetatraj}$
tends to $\frac{\partial h(t)}{\partial \theta_t}$, and since $\tilde
G_t$ is an unbiased estimate of $G_t$, the situation gets closer and
closer to an ordinary stochastic gradient descent if $\eta$ is small.
Presumably this happens whenever the learning rate is small enough for
$\theta$ not to change too much within a time range corresponding to a
``forgetting time'' of the dynamical system, although more work is needed
here.
}
% Consequently, for learning rates tending to $0$, the theory of stochastic
% gradient descent applies, and the trajectory of $\theta$ under this
% algorithm will asymptotically match the trajectory of real-time recurrent
% learning (based on maintaining the exact gradient $G(t)$) with the same
% learning rates, provided the dynamical system has enough
% ergodicity.

\subsection{Feeding the gradient estimate to an extended Kalman filter}

The Euclidean version of the NoBackTrack algorithm presented in
Algorithm~\ref{alg:eucl} is not enough to obtain good performance fast. Online
estimation often yields best results when using filters from the Kalman
family. We refer to \cite{Haykin_book,Jaeger_tutorial} for a discussion
of Kalman filtering applied to recurrent neural networks.

Kalman-based approaches rely on a covariance matrix
estimate $P(t)$ on $\theta$. After
observing $y(t)$, the parameter $\theta$ gets adjusted
via\footnote{Indeed, in standard Kalman filter notation, one has
$K_tR=P_t \transp{H_t}$, so that for the quadratic loss
$\ell=\frac12\transp{(\hat
y-y)}R^{-1}(\hat y-y)$ (log-loss of a Gaussian model with coraviance
matrix $R$), the Kalman update for $\theta$
is equivalent to $\theta\gets \theta - P(t)\transp{\frac{\partial
\ell_t}{\partial \theta}}$.}
\begin{equation}
\theta\gets \theta - P(t)\transp{\frac{\partial \ell_t}{\partial \theta}}
\end{equation}
where the derivative of the loss with respect to $\theta$ is
computed, as above, via the product of the derivative of the loss with
respect to the current state $h(t)$, and the derivative
$G(t)=\frac{\partial h(t)}{\partial \theta}$.

Maintaining a full covariance matrix on $\theta$ is usually too costly.
However, having a good approximation of $P(t)$ is not as critical as
having a good approximation of $\frac{\partial \ell_t}{\partial \theta}$.
Indeed, given an unbiased approximation of $\frac{\partial
\ell_t}{\partial \theta}$, \emph{any} symmetric positive definite matrix
$P(t)$ which changes slowly enough in time will yield an unbiased
trajectory for $\theta$.

Thus, we will use more aggressive matrix
reduction techniques on $P(t)$, such as block-diagonal (as in
\cite{Haykin_book}) or quasi-diagonal \cite{gradnn}
approximations. In our setting, the main point of using the covariance
matrix is to get both a sensible scaling of the learning rate for each
component of $\theta$, and reparametrization-invariance
properties \cite{gradnn}.

In Kalman filtering, in the case when the ``true'' underlying parameter $\theta$ in the extended Kalman
filter is constant, it is better to work with the inverse covariance
matrix $J(t)\deq P(t)^{-1}$, and the extended Kalman filter on $\theta$
can be rewritten as
\begin{align}
J(t)&\gets J(t-1)+\transp{\frac{\partial \hat y_t}{\partial
\theta}} I_t %\left(-\frac{\partial^2 \ell_t}{\partial \hat y_t^2}\right)
\frac{\partial \hat y_t}{\partial
\theta}
\\
\theta&\gets \theta-J(t)^{-1}\transp{\frac{\partial \ell_t}{\partial
\theta}}
\end{align}
where $\hat y_t$ is the prediction at time $t$, where both
$\frac{\partial \hat y_t}{\partial \theta}$ and $\frac{\partial
\ell_t}{\partial \theta}$ can be computed from $h(t)$ via the chain rule if
$G(t)=\frac{\partial h(t)}{\partial \theta}$ is known, and where $I_t$ is
the Fisher information matrix of $\hat y_t$ as a probability distribution
on $y_t$. (For exponential families this is just the Hessian
$-\frac{\partial^2 \ell_t}{\partial \hat y_t^2}$ of the loss with respect
to the prediction).
This is the so-called \emph{information filter}, because $J(t)$
approximates the Fisher information matrix on $\theta$ given the
observations up to time $t$. This is basically a natural gradient descent
on $\theta$.

This approach is summarized in Algorithm~\ref{alg:kalman}, which we
describe more loosely since matrix approximation schemes may depend on
the application.

Algorithm~\ref{alg:kalman} uses a decay factor $(1-\gamma_t)$ on the
inverse covariance matrices to limit the influence of old computations
made with outdated values of $\theta$. The factor $\gamma_t$ also
controls the effective learning rate of the algorithm, since, in line
with Kalman filtering, we have not included a learning rate for the
update of $\theta$ (namely, $\eta_t=1$): the step size is adapted via the
magnitude of $J$. For $\gamma_t=0$, $J$ grows linearly so that step size
is $O(1/t)$.

\newcommand{\matred}{\mathrm{MatrixReduce}}

\begin{algorithm}
\Params{$h(0)$ (initial state), $\theta_0$, $\phi_0$ (initial value of
the parameters), $0\leq \gamma_t< 1$ (covariance decay
parameter scheme), $\Lambda_\phi$ and $\Lambda_\theta$ (inverse covariance matrix of the prior on
the parameters)\;}
\Maintains{Same as Algorithm~\ref{alg:eucl}, plus a representation
of matrices $J_\theta$ and $J_\phi$ allowing for efficient inversion\;}
\Subroutines{A matrix reduction method $\matred(M)$ which only
evaluates a small number of entries of its argument $M$ and returns an
approximation of $M$ that can be inverted efficiently\; A routine
$\mathrm{FisherApprox}(\hat y_t,y_t)$ which returns either a positive
definite approximation of
the Fisher information matrix of $\hat y_t$ as a probability distribution
on $y_t$, or a positive definite approximation of the Hessian $-\frac{\partial^2
\ell_t}{\partial \hat y_t^2}$ of the loss with respect to the prediction.}

\BlankLine

\textbf{Initialization:} as in Algorithm~\ref{alg:eucl}, and
$J_\theta\gets 0$,
$J_\phi\gets 0$\;

\For{$t=0$ \KwTo end-of-time}{

\textbf{Observation step:} as in Algorithm~\ref{alg:eucl}.

\textbf{Update step:}
Compute approximate Fisher information matrix w.r.t.~$\hat y_t$:
\begin{equation}
I_t\gets \mathrm{FisherApprox}(\hat y_t,y_t)
\end{equation}

Compute derivative of prediction and of loss with respect to output
parameters, $\frac{\partial \hat y_t}{\partial \phi}$ and $\frac{\partial
\ell_t}{\partial \phi}$.
Update
inverse covariance matrix of output parameters $\phi$:
\begin{equation}
J_\phi \gets (1-\gamma_t) J_\phi+ \matred\left(\transp{\frac{\partial
\hat y_t}{\partial \phi}}I_t \,\frac{\partial
\hat y_t}{\partial \phi}\right)
\end{equation}
and
update output parameters:
\begin{equation}
\phi\gets \phi-(J_\phi+\Lambda_\phi)^{-1} \,\transp{\frac{\partial
\ell_t}{\partial \phi}}
\end{equation}

Compute derivative $\frac{\partial \hat y_t}{\partial h}$ of prediction with respect to current
state $h(t)$. Update inverse covariance matrix of internal parameters
$\theta$:
\begin{equation}
J_\theta\gets (1-\gamma_t) J_\theta + \matred\left(
\transp{\tilde G}\,\transp{\frac{\partial
\hat y_t}{\partial h}}I_t \,\frac{\partial
\hat y_t}{\partial h}\tilde G
\right)
\end{equation}
and update internal parameters $\theta$:
\begin{equation}
\theta\gets \theta-(J_\theta+\Lambda_\theta)^{-1} \delta\theta
\end{equation}
where $\delta\theta\deq  (Hv) \bar w -{\textstyle{\sum_i}} H_i w_i$ is
the update of $\theta$ from Algorithm~\ref{alg:eucl}.

\textbf{Reduction step:} Same as in Algorithm~\ref{alg:eucl}, but the
norms used to compute $\bar \rho$ and
$\rho_i$ are
derived from $J_\theta^{-1}$ (cf.\ Appendix~\ref{sec:invnorms}).

\textbf{Transition step:} Same as in Algorithm~\ref{alg:eucl}.

}

\caption{NoBackTrack algorithm, Kalman version.}
\label{alg:kalman}
\end{algorithm}

Moreover, we have included a regularization term
$\Lambda$ for matrix inversion; in the Bayesian interpretation of Kalman
filtering this corresponds to having a Gaussian prior on the parameters
with inverse covariance matrix $\Lambda$. This is important to avoid fast
divergence in the very first steps.

In practice we have used $\gamma_t=O(1/\sqrt{t})$ and $\Lambda=(\dim
h).\Id$.

The simplest and fastest way to approximate the Fisher matrix in
Algorithm~\ref{alg:kalman} is
the outer product approximation
(see discussion in
\cite{gradnn}), which we have used in the experiments below. Namely, we
simply use $I_t\gets \transp{\frac{\partial \ell_t}{\partial \hat y_t}}
\frac{\partial \ell_t}{\partial \hat y_t}$ so that the updates to
$J_\phi$ and $J_\theta$ simplify and become rank-one outer product updates
using the gradient of the loss, namely, $J_\theta\gets (1-\gamma_t)J_\theta
+ \transp{\frac{\partial \ell_t}{\partial \theta}}\frac{\partial
\ell_t}{\partial \theta}$ and likewise for $\phi$. Here the derivative
$\frac{\partial
\ell_t}{\partial \theta}$ is estimated from the current gradient
estimate $\tilde G$.

For the matrix reductions, we have used a
block-wise quasi-diagonal reduction as in \cite{gradnn}. This makes the
cost of handling the various matrices linear in the number of parameters.

\subsection{Examples}
\label{sec:examples}

Let us show how Algorithm~\ref{alg:eucl} works out
on explicit examples.

\new{
\paragraph{The importance of norm rescaling.}
Let us first consider a simple dynamical system which illustrates the
importance of rescaling the norms by $\bar\rho$ and $\rho_i$. Let
$0<\alpha<1$ and consider the system
\begin{equation}
h(t+1)=(1-\alpha) h(t)+\theta
\end{equation}
with both $h$ and $\theta$ in $\R^n$.
This quickly converges towards $\theta/\alpha$. We have $\partial
f/\partial h=(1-\alpha)\Id$ and $\partial f/\partial\theta=\Id$ and so
$\transp{\partial f_i/\partial\theta}=e_i$, the $i$-th basis vector. Then
the reduction and transition steps in Algorithm~\ref{alg:eucl}, if the
scalings $\rho$ are not used, amount to
\begin{align}
\bar v_{t+1} &= (1-\alpha)\left(\bar v_t+ {\textstyle\sum_i} \eps_i(t) e_i\right)
\\
\bar w_{t+1} &= \bar w_t + {\textstyle\sum_i} \eps_i(t) e_i
\end{align}
with the $\eps_i(t)$ independent at each step $t$.
The resulting estimate of $\partial h(t)/\partial \theta$ is
unbiased, but its variance grows linearly with time. Indeed, the dynamics
of $\bar v_t$ is stationary thanks to the factor $(1-\alpha)$, but the
dynamics of $\bar w_t$ is purely additive so that $w_t$ is just a
$d$-dimensional random walk. On the other hand, if rescaling by $\rho$ is
used, then both $\bar v$ and $\bar w$ get rescaled by $\sqrt{1-\alpha}$
at each step,\footnote{Proof: By induction one has $\bar v=\bar w$ after
the reduction step and $\bar v=(1-\alpha)\bar w$ after the transition
step, and $\bar\rho=1/\sqrt{1-\alpha}$.} so that their dynamics
becomes stationary and variance does not grow.

}%\new

\paragraph{Recurrent neural networks.}
The next example is a standard recurrent neural network (RNN).
The state of the system is the set of pre-activation values $h_i(t)$, and
the activities are $a_i(t)\deq \sigma(h_i(t))$ where
$\sigma$ is some activation function such as tanh or sigmoid. The recurrent dynamics of
$h$ is
\begin{equation}
	h_i(t+1)=\sum\limits_{j\to i}W_{ji} \,\sigma(h_j(t)) +
	\sum\limits_l r_{li} x_l(t)\\
	\label{eq:rnndyn}
\end{equation}
in which $h(t), h(t+1)\in \R^n$,
$(W_{ji})_{j\to i}$ are a set of weights defining a graph on 
$n$ nodes, and $(r_{li})_{(i,l)}$
are the input weights.\footnote{Biases are omitted; they can be treated by
the inclusion of an always-activated united $i_0$ with $a_{i_0}(t)\equiv
1$.} The parameter is $\theta=(W,r)$. We hereby
omit the output part of the network,\footnote{The experiments below use a
softmax output with output parameters $\phi$, see Section~\ref{sec:exp}.} as it is of no use to analyze
the estimation of $\partial h(t)/\partial \theta$.

(We have chosen the pre-activation values $h$, rather than the activities
$a=\sigma(h)$, as the state of the system.
This results in simpler expressions, especially for the input
weights $r$.)

Thus, the function $f$ defining the dynamical system for the variable $h$ is
\eqref{eq:rnndyn}. The derivatives of $f$ are immediately computed as
$\partial f_i/\partial W_{ji} = \sigma(h_j)$,
$\partial f_i/\partial r_{li} = x_l$, $\partial f_i/\partial h_j = W_{ji}
\,\sigma'(h_j)$, and all other derivatives are $0$.

Algorithm~\ref{alg:eucl} maintains, after the reduction
step, an approximation $\frac{\partial h(t)}{\partial \theta}\approx\bar
v(t)\transp{\bar w(t)}$. We can decompose $\bar w(t)=(\bar W(t),\bar
r(t))$ into the components corresponding to the internal
and input weights of the parameter $\theta=(W,r)$, so that
\begin{align}
	\frac{\partial h_i(t)}{\partial W_{kj}} &\approx \bar v_i(t) \bar
	W_{kj}(t)\\
	\frac{\partial h_i(t)}{\partial r_{lj}} &\approx \bar v_i(t) \bar
	r_{lj}(t).
\end{align}

By plugging the values of the partial derivatives of $f$ into
Algorithm~\ref{alg:eucl}, we find the following update equations for the
value of $\bar v$, $\bar W$ and $\bar r$ right after the reduction step:
% 
% Following the reduction trick exposed in Proposition~\ref{prop:rk1},
% we get the following update rules for the rank-one approximation of the
% differential state of the network\NDY{I think
% using $\rho$ for something that is close to, but not the $\rho$ of the
% algorithm will be confusing.}
\begin{align}
	\label{eq:rnnvbar}
	\bar v_i(t+1) &= \bar\rho \sum\limits_{j\to i}W_{ji}
	\,\sigma'(h_j(t))\,\bar v_j(t) + \eps_i\rho_i\\
	\label{eq:rnnWbar}
	\bar W_{kj}(t+0) &= \frac{\bar W_{kj}(t)}{\bar\rho} + \eps_j \frac{\sigma(h_k(t))}{\rho_j}\\
	\label{eq:rnnrbar}
	\bar r_{lj}(t+1) &= \frac{\bar r_{lj}(t)}{\bar\rho} + \eps_j
	\frac{x_l(t)}{\rho_j}
\end{align}
where the $\eps_j$ are independent symmetric binary random variables, taking values 
$\pm1$ with probability $\frac{1}{2}$. Any non-zero
choice of $\rho_j$ leads to an unbiased estimation, though the values are
to be optimized as mentioned above.

Applying this update has the same algorithmic cost as implementing one
step \eqref{eq:rnndyn} of the recurrent network itself.

\paragraph{Leaky recurrent neural networks.}
To capture long-term dependencies, in the experiments below we also use a
\emph{leaky} RNN, obtained via
the addition of a direct feedback
term:
\begin{equation}
	h_i(t+1)=\alpha_i h_i(t)+ \sum_l r_{li} x_l(t)+\sum_j W_{ji} a_j(t)
 \quad a_j(t)\deq \sigma(h_j(t))
\end{equation}
with $\alpha_i \in \left[0;1\right]$ for all $i$. (See
\cite{Jaeger_tutorial} for similar
models.) This feedback term reduces the impact
of the vanishing gradient issue and keeps a longer memory of past inputs.

This only changes the derivative of $f_i$ with respect to
$h_j$, which becomes $\partial f_i/\partial h_j=W_{ji}\sigma'(h_j) + \alpha_i
\delta_{ij}$. Consequently the update rules \eqref{eq:rnnWbar}--\eqref{eq:rnnrbar} for
$\bar W$ and $\bar r$ are unchanged, while 
the update of $\bar v$ becomes 
\begin{equation}
\bar v_i(t+1) = \bar\rho \alpha_i \bar v_i(t)+\bar\rho \sum\limits_{j\to
i}W_{ji}\,
\sigma'(h_j(t))\,\bar v_j(t)  + \eps_i\rho_i
\end{equation}

\paragraph{Multilayer recurrent neural networks.}
Let us now treat the case of a multilayer recurrent neural
network with dynamics
\begin{align}
	h^{(1)}(t+1) &= f^{(1)}(x(t),h^{(1)}(t),\theta_1)\\
	h^{(2)}(t+1) &= f^{(2)}(x(t),h^{(1)}(t+1),h^{(2)}(t),\theta_2)\\
	&\vdots\\
  h^{(n)}(t+1) &= f^{(n)}(x(t),h^{(n-1)}(t+1),h^{(n)}(t),\theta_n)
\end{align}
where each layer $h^{(i)}$ and $f^{(i)}$ define an RNN as in \eqref{eq:rnndyn} above. Directly
applying the rank-one approximation to the function $f=(f^{(1)},f^{(2)},\ldots,f^{(n)})$ would
be cumbersome: since the activity of a neuron of the $i$-th layer at time
$t+1$ depends on all parameters 
from the previous $i-1$ layers, the derivative $\partial f/\partial
\theta$ is not sparse.

To cope with this, a natural approach is to treat the dynamics in a
``rolling'' fashion and apply the rank-one approximation at each layer in
turn.
Formally, this amounts to defining the
following model
\begin{align}
	\tilde{h}^{(1)}(nt+1) &= f^{(1)}(\tilde{x}(nt),\tilde{h}^{(1)}(nt),\theta_1)\\
	\tilde{h}^{(2)}(nt+2) &= f^{(2)}(\tilde{x}(nt+1),\tilde{h}^{(2)}(nt+1),\theta_2)\\
	&\vdots\\
	\tilde{h}^{(n)}(nt+n) &=
	f^{(n)}(\tilde{x}(nt+n-1),\tilde{h}^{(n)}(nt+n-1),\theta_n)
\end{align}
with $\tilde{x}(t) \deq x(\lfloor t/n\rfloor)$, and where states
not explicitly appearing in these equations stay unchanged
($h^{(i)}(nt+j)=h^{(i)}(nt+j-1)$ for $i\neq j$).
Thus, the transition function
explicitly depends on time (more precisely, on time modulo the number
of layers), and is sparse at each step. Indeed, at each step, applying the transition
function amounts to applying one of the $f^{(i)}$ to the corresponding layer,
and leaving the other layers unchanged. Thus the derivative of $f^{(i)}$ with
respect to any $\theta_j$, $j\neq
i$, is zero; this leaves only the gradient of $f^{(i)}$ wrt $\theta_i$ to be
dealt with, and Algorithm~\ref{alg:eucl} or \ref{alg:kalman} can be applied at little cost.

\subsection{Extensions}
\label{sec:extensions}

\paragraph{Rank-$K$ reductions.}
A first obvious extension is to use higher-rank
reductions. The simplest way to achieve this is to take several
independent random rank-one $\bar v_k \transp{\bar w_k}$ reductions in \eqref{eq:Atilde} and average
them. Note that $w_i$ (Algorithm~\ref{alg:eucl}) has to be evaluated only
once in this case. It might be slightly more efficient to first split the
parameter components into $K$ blocks (e.g., at random) so that the $k$-th term
$\transp{\bar w_k}$ only involves parameters from the $k$-th block:
indeed, applying the evolution equation for $G$ preserves this structure
so this requires less memory for storage of the $\bar w_k$.

% \NDY{Not sure whether to keep this, WDYT?} Another extension would be to keep a diagonal plus rank-one
% approximation. This amounts to maintaining $\bar v\transp{\bar w}+\sum_i
% e_i\transp{w_i}$ and applying the reduction step \emph{after} the
% transition step; namely, after the transition step the gradient is
% $\left(\frac{\partial f}{\partial h} \bar v\right)\transp{\bar w}+\sum_i 
% \left(\frac{\partial f}{\partial h_i} \transp{w_i}+e_i \transp{\frac{\partial
% f_i}{\partial \theta}}\right)$, where usually $\frac{\partial f}{\partial h_i}$
% is sparse. One can then separate the $e_i$ and non-$e_i$ components of
% $\frac{\partial f}{\partial h_i}$, keep the $e_i$ component, and apply
% the rank-one trick to merge the non-$e_i$ components into $\bar v$ and
% $\bar w$. This is especially relevant if some components of the system are
% almost decoupled, or for slowly evolving systems where $\frac{\partial
% f}{\partial h}$ is close to the identity.

\newcommand{\deltat}{\ensuremath{\hspace{0.05em}\delta\hspace{-.06em}t\hspace{0.05em}}}

\paragraph{Algorithms similar to RTRL.} Other algorithms have been proposed that have the same structure and
shortcomings as real-time recurrent learning, for instance, the online EM
algorithm for hidden Markov models from \cite{Cappe_onlineHMM}. In
principle, the approach presented here can be extended to such settings.

\paragraph{Continuous-time systems.} Another extension concerns continuous-time dynamical systems
\begin{equation}
\frac{\d h(t)}{\d t}=F(h(t),x(t),\theta)
\end{equation}
which can be discretized as $h(t+\deltat)=h(t)+\deltat
F(h(t),x(t),\theta)$. Thus this is analogous to the discrete-time case
via $f=\Id+\deltat F$, and Algorithm~\ref{alg:eucl} may be applied to
this discretization.

When performing the rank-one reduction
\eqref{eq:Atilde}, the scaling by $\rho_i=\sqrt{\norm{w_i}/\norm{v_i}}$
is important in this case: it ensures that both $\bar v$ and $\bar w$ change by
$O(\sqrt{\deltat})$ times a random quantity at each step. This is the
expected correct scaling for a continuous-time stochastic evolution
equation, corresponding to the increment of a Wiener process during a
time interval $\deltat$. (Without scaling by $\rho_i$, there will be no well-defined
limit as $\deltat\to 0$, because $\bar v$ would change by $O(1)$ at each
step $t\gets t+\deltat$, while $\bar w$ would evolve by $\deltat$ times a
centered random quantity so that it would be constant in the limit.)
Further work is needed to study this continuous-time limit.

\section{Experiments}
\label{sec:exp}

\newcommand{\Al}{\mathcal{A}}

We report here a series of small-scale experiments on text prediction
tasks. The experiments focus on two questions: First, does learning using
the rank-one approximation $\tilde G$ accurately reflect learning based
on the actual gradient $G$ computed exactly via RTRL, or is the noise
introduced in this method detrimental to learning? Second, how does this
approach compare to truncated backpropagation through time?

We used the RNN or leaky RNN models described above to predict a
sequence of characters $y(t)$ in a finite alphabet $\Al$, given the
past observations $x(s)=y(s)$ for $1\leq s\leq t-1$. At each time, the network outputs a
probability distribution on the next character $z$; explicitly,
the output at time $t$ is $\hat y(t)\in \R^\Al$ defined by
\begin{equation}
\hat y(t)_z \deq \phi_z+ \sum_i \phi_{iz} a_i(t)
\end{equation}
for each $z\in \Al$, with parameters
$\phi=(\phi_z,\phi_{iz})_{i,z}$.
The output $\hat y=(\hat y_y)_{y\in\Al}$ defines a probability distribution on $\Al$ via a
softmax 
$p_{\hat y}(y)\deq \frac{e^{\hat y_y}}{\sum_{z\in Al} e^{\hat y_z}}$, and the
loss function is the log-loss on prediction of the next character, $\ell_t\deq -\log_2
p_{\hat y(t)}(y(t))$. The internal and output parameters $\theta$ and $\phi$ are trained according
to Algorithms~\ref{alg:eucl} and \ref{alg:kalman}.

We used three datasets. The first is a ``text'' representing synthetic
music notation with several syntactic, rhythmic and harmonic constraints
(Example~3 from \cite{pcnn}). The data was a file of length $\approx
10^5$ characters, after which the signal cycled over the same file.  The
second dataset is the classical $a^nb^n$ example,
synthesized by repeatedly picking an integer $n$ at random in some
interval, then outputting a series of $n$ $a$'s followed by a
line break, then $n$ $b$'s and another line break. This model
tests the ability of a learning algorithm to learn
precise timing and time dependencies.
The
third example is the full set of Shakespeare's works, obtained from Project
Gutenberg.\footnote{\url{www.gutenberg.org}} The file is roughly $5.10^6$
characters long.

\newcommand{\gzip}{\texttt{gzip}}

The benchmarks included are \gzip, a standard non-online
compression algorithm, and context tree weighting (CTW)
\cite{elyaniv_markov2004}, a more advanced
online text compression algorithm, as well as the actual entropy rate of
the generative model for synthetic music and $a^nb^n$.

The code used in the experiments is available at
\url{http://www.yann-ollivier.org/rech/code/nobacktrack/code_nobacktrack_exp.tar.gz}

\paragraph{Euclidean NoBackTrack.}
We first study whether the low rank approximation in the \emph{Euclidean}
version of NoBackTrack impacts the gradient descent.
For this first set of
experiments, we use a fully connected RNN with $20$ units, as described
above, on the synthetic music example. We compared RTRL, Euclidean
rank-one NoBackTrack, and Euclidean NoBackTrack using rank-two and
rank-ten reductions (obtained by averaging two or ten independent
rank-one reductions, as discussion in Section~\ref{sec:extensions}).

\begin{figure}
\begin{center}
\includegraphics[width=.75\textwidth]{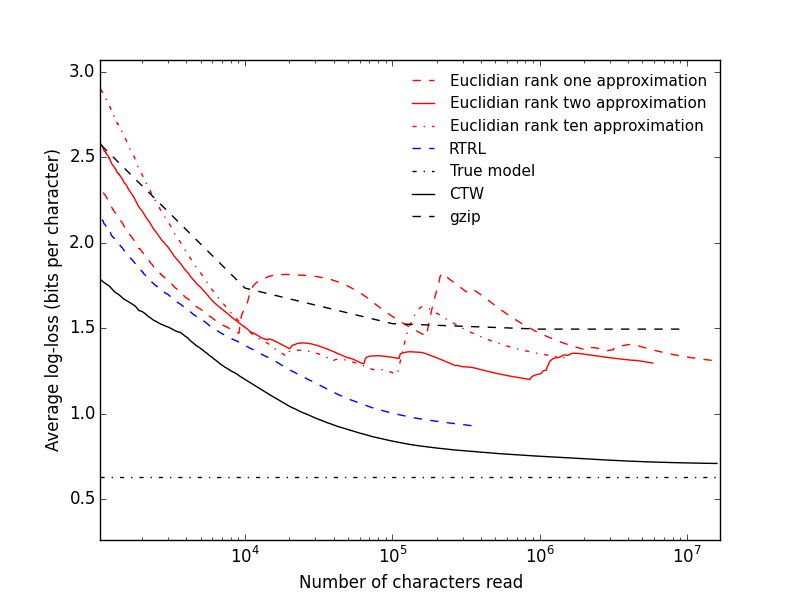}
\end{center}
\caption{Average log-loss (bits per character) on synthetic music as a function of the
number of characters read, for an RNN with 20 units trained with the Euclidean
version of the NoBackTrack algorithm for different rank values and RTRL, with
learning rate $\eta_t=1/\sqrt{t}$, benchmarked against the true model
entropy rate, \gzip, and CTW.}
\label{fig:eucl}
\end{figure}

\begin{figure}
\begin{center}
\includegraphics[width=.75\textwidth]{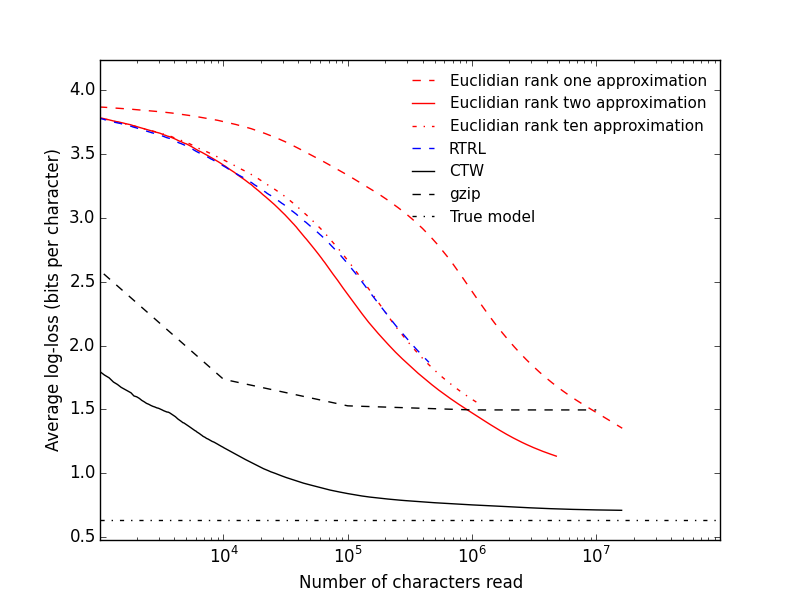}
\end{center}
\caption{Average log-loss (bits per character) on synthetic music as a function of the
number of characters read, for an RNN with 20 units trained with the Euclidean
version of the NoBackTrack algorithm for different rank values, and with RTRL, with
learning rate $\eta_t=0.03/\sqrt{t}$, benchmarked against the true model
entropy rate, \gzip, and CTW.}
\label{fig:eucl_low}
\end{figure}

The results are summed up in Figure~\ref{fig:eucl} and Figure~\ref{fig:eucl_low}. All the models were
trained using the same learning rate $\eta_t=1/\sqrt{t}$ for
Figure~\ref{fig:eucl} and 
$\eta_t=0.03/\sqrt{t}$ for Figure~\ref{fig:eucl_low}.

The various algorithms were run for the
same amount of time. This is reflected in the different curve lengths for
the different algorithms; in particular, the curve for RTRL is much
shorter, reflecting its higher computational cost. (Note the log scale on
the $t$ axis: RTRL is roughly $20$ times slower with $20$ units.)


The impact of stochasticity of the low-rank approximation when using
large learning rates is
highlighted on Figure~\ref{fig:eucl}: Euclidean NoBackTrack with a large
learning rate displays instabilities, even when increasing the rank of
the approximation.

Smaller learning rates allow the algorithm to cope with this, as the
noise in the gradients is averaged out over longer time spans. This is
illustrated in Figure~\ref{fig:eucl_low}, in which the trajectories of
Euclidean NoBackTrack track those of RTRL closely even with a
rank-two approximation.

\paragraph{Kalman NoBackTrack.}
Next, we report the results of the Kalman version of NoBackTrack on the same
experimental setup. A quasi-diagonal outer product (QDOP) approximation
\cite{gradnn}
of the full Kalman inverse covariance matrix
is used, to keep complexity low.

We compare the low-rank approximations to RTRL. To make the comparison
clear, for RTRL we also use a quasi-diagonal (QDOP) approximation of the Kalman
filtering algorithm on top of the exact gradient computed by RTRL.

Learning rates
were set to $1$ and all algorithms were run for the same amount of time.
\begin{figure}
\begin{center}
\includegraphics[width=.75\textwidth]{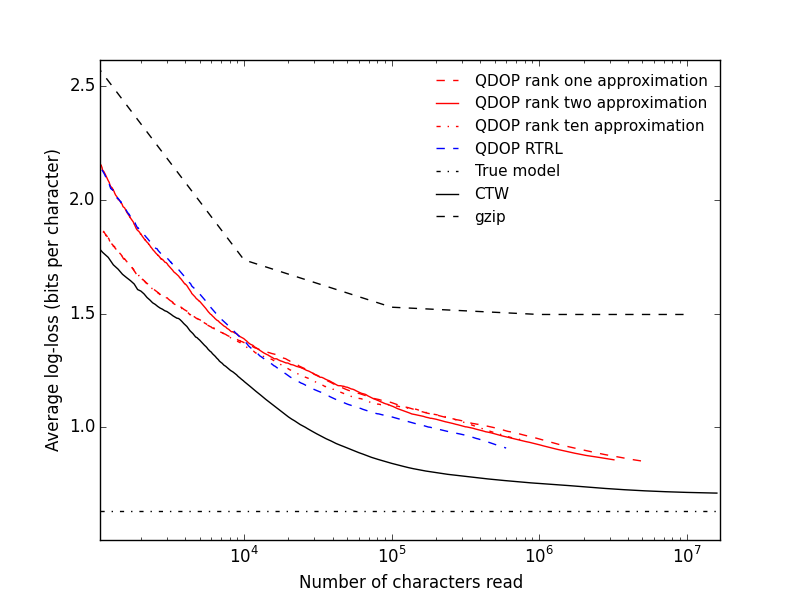}
\end{center}
\caption{Average log-loss (bits per character) on synthetic music as a function of the
number of characters read, for an RNN with 20 units trained with the
Kalman/QDOP
version of the NoBackTrack algorithm for different rank values and
Kalman/QDOP
RTRL, benchmarked against the true model
entropy rate, \gzip, and CTW.}
\label{fig:qdop}
\end{figure}

The use of the QDOP-approximated Kalman inverse covariance appears to
fully fix the unstable behaviour. Overall,
low-rank approximations appear to be roughly on par with QDOP RTRL.
There is no obvious gain, on this particular example, in using
higher-rank approximations.

Still, on this particular task and with this particular network size,
none of the RNN algorithms (including BPTT reported below) match the
performance of Context Tree Weighting.  RNNs beat CTW
on this task if trained using a non-online,
Riemannian gradient descent \cite{pcnn} (analogous to using the Kalman
inverse covariance). So this is arguably an effect of imperfect online
RNN training.

\paragraph{Kalman NoBackTrack and truncated BPTT.}
Our next set of experiments aims at comparing Kalman NoBackTrack to truncated
BPTT, with truncation\footnote{In the version of BPTT used here, the
algorithm does not backtrack by $T$ steps at every time step $t$; rather,
it waits for $T$ steps between $t$ and $t+T$, then backtracks by $T$
steps and collects all gradients in this interval. Otherwise, truncated
BPTT would be $T$ times slower, which was unacceptable for our
experiments.} parameter $T=15$. As BPTT truncates the full
gradient by removing dependencies at distances longer than
the truncation parameter, we expect Kalman NoBackTrack to learn better models
on datasets presenting long term correlations.

\begin{figure}
\begin{center}
\includegraphics[width=.75\textwidth]{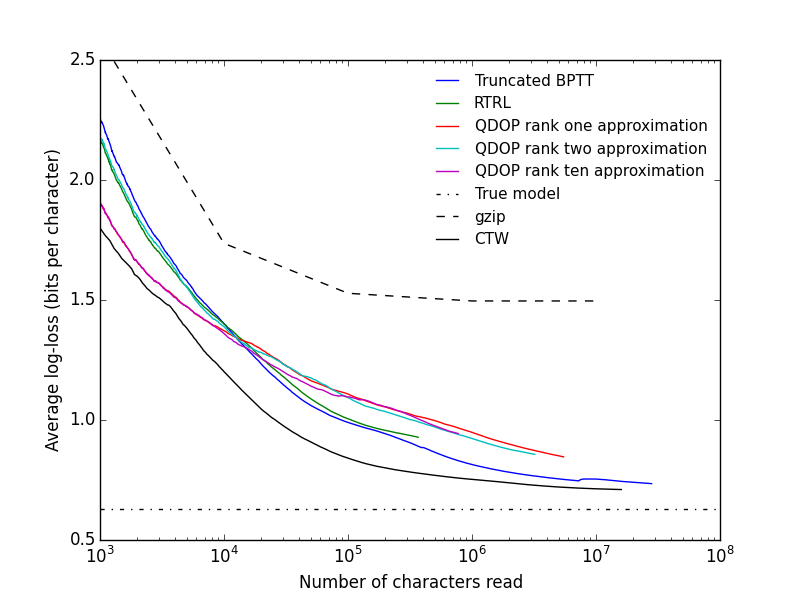}
\end{center}
\caption{Average log-loss (bits per character) on synthetic music as a function of the
number of characters read, for an RNN with 20 units trained with the
Kalman/QDOP
version of the NoBackTrack algorithm for different rank values, Euclidean RTRL, and truncated
BPTT, benchmarked against the true model
entropy rate, \gzip, and CTW.}
\label{fig:rk_comp}
\end{figure}

The two algorithms are first compared on the synthetic music dataset,
with the same experimental setup as above, for the same amount of time,
with a learning rate $\eta_t=1/\sqrt{t}$ for truncated BPTT and
$\gamma_t=1/\sqrt{t}$ for Kalman NoBackTrack.\footnote{These learning
rates have different meanings for Kalman NoBackTrack and truncated BPTT,
and are not directly comparable.}  The results are shown in
Figure~\ref{fig:rk_comp}.

On this example, truncated BPTT perfoms better than Kalman NoBackTrack,
even though the two algorithms display broadly comparable performance.
Noticeably, RTRL and truncated BPTT are roughly on par here, with
truncated BPTT
slightly outperforming RTRL in the end: apparently, 
maintaining
long term dependencies in gradient calculations does not improve
learning in this synthetic music example.

% Overall, the observations made above still hold. The relative efficiency of
% algorithms is maintained at the end of learning. There again truncated BPTT
% performs better than RTRL, and it is unclear wether truncating long term
% gradient terms might hurt the learning, as the example is complex enough for
% having huge improvement by only learning short term dependencies.

\bigskip

Next, to compare NoBackTrack and truncated BPTT on their specific
ability to learn precise middle and long term dependencies, we present
experiments on the $a^nb^n$ example. This will clearly illustrate the
biased nature of the gradients computed by truncated BPTT.

\begin{figure}
\begin{center}
\includegraphics[width=.75\textwidth]{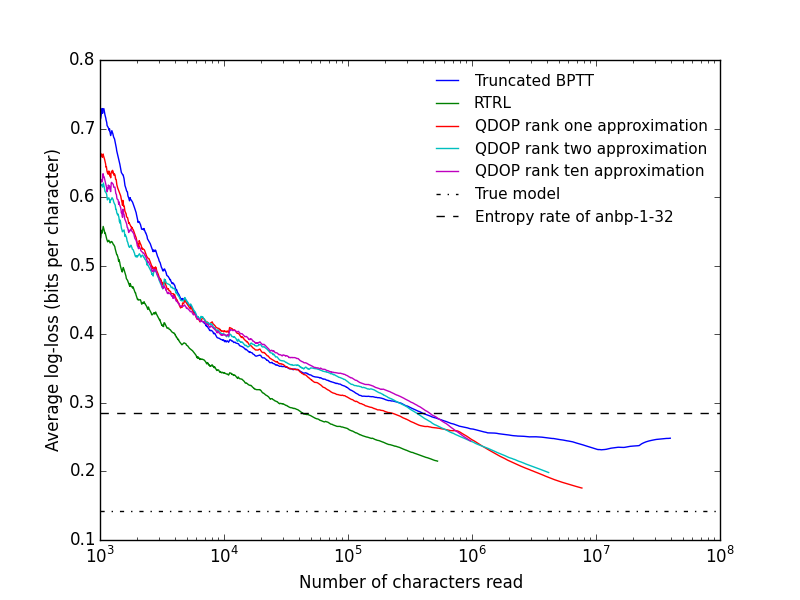}
\end{center}
\caption{Average log-loss (bits per character) on the $a^nb^n_{[1,32]}$
dataset, as a function of the
number of characters read, for a leaky RNN with 20 units, trained with a
Kalman/QDOP version of the NoBackTrack algorithm, RTRL, and BPTT.}
\label{fig:lrnn_rk_comp}
\end{figure}

The $a^nb^n_{[k;l]}$ dataset is synthesized by sequentially picking a number $n$ between $k$
and $l$ uniformly at random, then outputting a series of $n$ $a$'s followed by a
line break, then $n$ $b$'s and another line break. The true entropy rate is
$\frac {\log_2 (l-k+1)}{l+k+2}$ in this example.\footnote{Indeed, $\log_2(l-k+1)$
bits are needed to encode the value of $n$ in each $a^nb^n$ block
(this is the entropy of a uniform distribution on
$\{k,\ldots,l\}$), and the average value of $n$ is $(k+l)/2$ so that the
average length of an $a^nb^n$ block, including
the two newline symbols, is $2\times (l+k)/2+2$.}
A roughly $10^6$ character long input sequence was
synthesized, using $[k;l]=[1;32]$.

As standard RNN models do not seem to be able to deal with this example,
whatever the training algorithm, we used a leaky RNN\footnote{The
parameter $\alpha$
	of the LRNN can be learned, but this sometimes produces
	numerical instabilities unless cumbersome changes of variables are
	introduced.
	We just initialized $\alpha$ to a
	random value separately for each unit and kept it fixed.}  as presented in
Section~\ref{sec:examples}, again with $20$ fully
connected units.
All the algorithms used a learning rate of $1/\sqrt{t}$. The results are
reported on Figure~\ref{fig:lrnn_rk_comp}, which also includes the
entropy rate of the exact $a^nb^n$ model and the (twice larger) entropy
rate of an $a^nb^p$ model with independent $n$ and $p$.

Kalman NoBackTrack clearly outperforms
truncated BPTT on this dataset. This was to be expected, as the typical
time range of the temporal dependencies exceeds the truncation range for
BPTT, so that the approximated gradients computed by truncated BPTT are significantly
biased.

Keeping track of the long term dependencies is key here, and RTRL
outperforms all the algorithms epochwise, though it is still penalized by its high
complexity. Truncated BPTT is unable to learn the full dependencies between $a$'s
and $b$'s, and ends up closer to the entropy of an $a^nb^p$ model with
independent values of $n$ and $p$ (presumably, it still manages to learn
the $a^nb^n$ blocks where $n$ is short). At some point the learning curve
of truncated BPTT appears not to decrease anymore and even goes slightly
up, which is consistent with a biased gradient estimate.

On the other hand, Kalman NoBackTrack seems to be mostly successful in
learning the dependencies. This is confirmed by visual inspection of the
output of the learned model. The small remaining gap between the true model
and the learned model could be related to incomplete training, or to an
imperfect modelling of the exact uniform law for $n\in[k;l]$.

\bigskip

\begin{figure}
\begin{center}
\includegraphics[width=.75\textwidth]{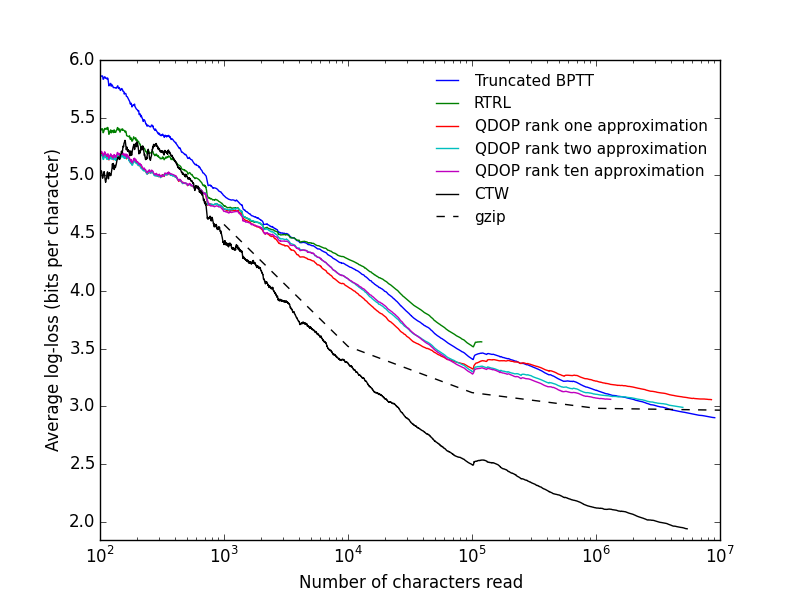}
\end{center}
\caption{Average log-loss (bits per character) on Shakespeare's works as a function of the
number of characters read, for an RNN with 20 units trained with the QDOP
version of the NoBackTrack algorithm for different rank values, Euclidean RTRL and truncated
BPTT, benchmarked against \gzip{} and CTW.}
\label{fig:rk_comp_shak}
\end{figure}

Finally, we report performance of truncated BPTT and Kalman NoBackTrack on
Shakespeare's works. The same $20$-unit RNN model is used, again with all
algorithms run for the same amount of time using the same learning rate 
$1/\sqrt{t}$. The curves obtained are displayed in
Figure~\ref{fig:rk_comp_shak}.

On this example, RTRL, truncated BPTT, and Kalman NoBackTrack with various
ranks all have a similar performance; it is not clear whether the
differences on Figure~\ref{fig:rk_comp_shak} are statistically
significant. This proves, once more, that the stochasticity and rank
reduction inherent to NoBackTrack are not detrimental to learning, and
allow it to keep up with exact gradient algorithms.

All RNN algorithms have a significantly worse performance than CTW on
this example,
thus proving that a $20$-unit RNN does not accurately model Shakespeare's
works.

\bigskip

\new{
\paragraph{Conclusion.}
We have introduced an algorithm that computes a stochastic, provably unbiased
estimate of the derivative of the current state of a dynamical system with
respect to its parameters, in a fully online fashion. For recurrent neural
networks, the computational cost of this algorithm is comparable to that
of running the network itself. Previously known algorithms were either
not fully online
or had a significantly higher computational cost.

In our experiments, this algorithm appears as a practical alternative to
truncated backpropagation through time,
especially in its Kalman version, while the Euclidean
version requires smaller learning rates. The (unbiased) noise and rank
reduction introduced in the
gradient approximation do not appear to prevent learning. The interest
of NoBackTrack with respect to
truncated BPTT depends on the situation at hand, especially on the scale of
time dependencies in the data (which results in biased gradient estimates
for BPTT), and on whether the storage of past states
and past data required by
truncated BPTT is acceptable or not.
}

\new{
\paragraph{Acknowledgments.} The authors would like to thank Hugo
Larochelle for his helpful questions that resulted in several
clarifications of the text.}

%TODO: should also work reinforcement learning (with Bellman instead of
%backprop through time). And also in control theory as well, since the
%backward equations for optimal control have the same structure as
%backprop through time.

% \begin{prop}
% At each step of Algorithm~\ref{alg:eucl}, the quantity $\tilde G_t\deq
% \bar v\transp{\bar w}+\sum_i e_i \transp{w_i}$ satisfies
% \begin{equation}
% \E \tilde G_t=\frac{\partial h(t)}{\partial \theta}
% \end{equation}
% where the derivative is taken along the actual trajectory of parameters
% $\theta_t$.
% \end{prop}

%TODO: thanks: Rémi Peyre, Hugo Larochelle

{\small
\appendix

\section{Variance of the rank-one trick}
\label{app:varproof}

Keep the notation of Proposition~\ref{prop:rk1} and let $\norm{\cdot}$ be
a Euclidean norm on the vector space in which the $v_i$ and $w_i$ live.

\newcommand{\HSnorm}[1]{\norm{#1}_{\mathrm{HS}}}
\newcommand{\HSscal}[2]{\scal{#1}{#2}_{\mathrm{HS}}}

To measure the variance of $\tilde A$ we use the Hilbert--Schmidt norm
$\HSnorm{\tilde A}^2\deq\Tr(\transp{\tilde A\,}\tilde A)$. This norm satisfies
$\HSnorm{v\transp{w}}=\norm{v}\norm{w}$, and
$\HSscal{v_1\transp{w_1}}{v_2\transp{w_2}}=\scal{v_1}{v_2}\scal{w_1}{w_2}$
for the associated
scalar product.

Let us evaluate the variance of $\tilde A$ in this norm. Since $\Var
\tilde A= \E
\HSnorm{\tilde A}^2-\HSnorm{\E \tilde A}^2$ and $\E \tilde
A=A$ is fixed, it is enough to evaluate the second moment $\E
\HSnorm{\tilde A}^2$.

We claim that
\begin{equation}
\label{eq:secondmom}
\E \HSnorm{\tilde A}^2=(\sum_i \norm{v_i}^2)(\sum_j
\norm{w_j}^2)+2\sum_i\sum_{j\neq i}\scal{v_i}{v_j} \scal{w_i}{w_j}
\end{equation}
Indeed,
$\tilde A=\sum_{ij} \eps_i\eps_j v_i\transp{w_j}$ so, by bilinearity of
the Hilbert--Schmidt scalar product,
\begin{align}
\E \HSnorm{\tilde A}^2 &=\E\HSscal{\tilde A}{\tilde A}
=\E
\sum_{ijkl} \eps_i\eps_j\eps_k\eps_l \scal{v_i}{v_j} \scal{w_k}{w_l}
\end{align}
Since $\E \eps_i=0$ and $\E (\eps_i\eps_j)=0$ for $i\neq j$, the only cases to consider
are:
\begin{enumerate}
\item $i=j$ and $k=l$ and $i\neq k$: contribution $\sum_i \sum_{k\neq i}
\norm{v_i}^2 \norm{w_k}^2$
\item $i=k$ and $j=l$ and $i\neq j$: contribution $\sum_i \sum_{j\neq i}
\scal{v_i}{v_j}\scal{w_i}{w_j}$
\item $i=l$ and $j=k$ and $i\neq j$: same contribution as the previous one
\item $i=j=k=l$: contribution $\sum_i \norm{v_i}^2 \norm{w_i}^2$
\item all other cases contribute $0$.
\end{enumerate}
The first and fourth contributions add up to $(\sum_i \norm{v_i}^2)(\sum_k
\norm{w_k}^2)$. This proves \eqref{eq:secondmom}.

Let us minimize variance over the degrees of freedom given by
$v_i\transp{w_i}=(\rho_i v_i)\transp{(w_i/\rho_i)}$.
$\rho_i$ does not change the last contribution to $\E \HSnorm{\tilde A}^2$
in \eqref{eq:secondmom},
neither does it change the expectation $\E\tilde A=A$, so to minimize the variance we
only have to minimize the first term $(\sum_i \norm{v_i}^2)(\sum_k
\norm{w_k}^2)$. Applying the scaling, this term becomes
\begin{equation}
(\sum_i \norm{v_i}^2 \rho_i^2)(\sum_k
\norm{w_k}^2/\rho_k^2)
\end{equation}
and, by differentiation with respect to a single $\rho_i$, one checks that this is minimal for
\begin{equation}
\rho_i \propto \sqrt{\norm{w_i}/\norm{v_i}}
\end{equation}
(mutliplying all $\rho_i$'s by a common factor does not change
the result). So, after optimal scaling,
\begin{equation}
\tilde A=
\left(\sum_i \eps_i v_i \sqrt{\norm{w_i}/\norm{v_i}}\right)
\otimes
\left(\sum_i \eps_i w_i \sqrt{\norm{v_i}/\norm{w_i}}\right)
\end{equation}
Consequently, after scaling,
the first term in the variance of $\tilde A$ in \eqref{eq:secondmom}
becomes $(\sum_i \norm{v_i}\norm{w_i})^2$.
The second term in \eqref{eq:secondmom} does not change.

Thus, after optimal scaling we find
\begin{equation}
\E \HSnorm{\tilde A}^2=\left(\sum_i \norm{v_i}\norm{w_i}\right)^2
+2\sum_i\sum_{j\neq i}\scal{v_i}{v_j} \scal{w_i}{w_j}
\end{equation}
To obtain the variance of $\tilde A$, we just subtract the square norm of
$\E\tilde A=A$, which is
\begin{align}
\HSnorm{A}^2 &= \HSnorm{\sum_i v_i\transp{w_i}}^2
= \sum_i \HSnorm{v_i \transp{w_i}}^2 + \sum_i\sum_{j\neq i}
\HSscal{v_i \transp{w_i}}{v_j \transp{w_j}}
\\\intertext{(by bilinearity of the Hilbert--Schmidt scalar product)}
&=\sum_i \norm{v_i}^2\norm{w_i}^2+\sum_i\sum_{j\neq i}\scal{v_i}{v_j}
\scal{w_i}{w_j}
\end{align}

This yields, after optimal scaling,
\begin{align}
\Var \tilde A&= \left(\sum_i \norm{v_i}\norm{w_i}\right)^2-\sum_i
\norm{v_i}^2\norm{w_i}^2+\sum_i\sum_{j\neq i}\scal{v_i}{v_j}
\scal{w_i}{w_j}
\\&=\sum_i \sum_{j\neq i}
\norm{v_i}\norm{v_j}\norm{w_i}\norm{w_j}+\scal{v_i}{v_j}
\scal{w_i}{w_j}
\end{align}

% In this computation we have not supposed that the $w_i$ are orthogonal.
% In case the $w_i$ are orthogonal we just have
% \begin{equation}
% \E\norm{\tilde G}^2=(\sum_i \norm{v_i}\norm{w_i})^2
% \end{equation}
% and
% \begin{equation}
% \Var \tilde G=(\sum_i \norm{v_i}\norm{w_i})^2-\sum
% \norm{v_i}^2\norm{w_i}^2
% \end{equation}

\new{
\section{Invariant norms derived from the
Kalman covariance}
\label{sec:invnorms}

Algorithm~\ref{alg:kalman} is built to offer invariance properties (a
Kalman filter over a variable $\theta$ is invariant by affine
reparameterization of $\theta$, for instance). However, this only holds
if the norms $\norm{\bar v}$, $\norm{\bar w}$, $\norm{v_i}$,
$\norm{w_i}$, used to compute the scaling factors 
$\bar \rho= \sqrt{\norm{\bar w}/\norm{\bar v}}$ and
$\rho_i=\sqrt{\norm{w_i}/\norm{e_i}}$,
are themselves reparameterization-invariant.

This can be achieved if we decide to choose the scalings $\rho$ as to
minimize the variance of $\tilde G$ computed in the (Mahalanobis) norm
defined by the covariance matrix of $\theta$ and of $h$ appearing in the
Kalman filter.

Let $C_\theta$ be the covariance matrix of $\theta$ obtained in the
Kalman filter; in Algorithm~\ref{alg:kalman}, $C_\theta$ is approximated
by $C_\theta\approx J_\theta^{-1}$.

Any linear form on $\theta$, such as $\bar w$ and $w_i$, can be given a
norm by
\begin{equation}
\norm{\bar w}^2\deq \transp{\bar w}C_\theta \bar w \approx \transp{\bar
w} J_\theta^{-1} \bar w
\end{equation}
and likewise for $w_i$. This norm is invariant under
$\theta$-reparameterization.

Given the covariance $C_\theta$ of $\theta$ and the dependency
$G=\frac{\partial h}{\partial \theta}$ of $h$ with respect to $\theta$,
the covariance of $h$ is
\begin{equation}
C_h\deq GC_\theta\transp{G}
\end{equation}
and its inverse $J_h\deq C_h^{-1}$ can be used to define a norm for a tangent vector $v$ at state $h$
via
\begin{equation}
\norm{v}^2\deq \transp{v} J_h v
\end{equation}
which is also reparametrization-invariant.
(We use $J_\theta^{-1}$ for the norm of $w$ and $J_h$ for the norm of $v$
because $v$ is a tangent vector (covariant) at point $h$, while $w$ is a
linear form (contravariant) at point $\theta$.)

However, handling of full covariance matrices would be too costly. In
Algorithm~\ref{alg:kalman}, the inverse covariance $J_\theta$ of $\theta$
is already an approximation (diagonal, quasi-diagonal...) via
$\matred$. Moreover, here we only have access to an approximation $\tilde
G$ of $G$. Thus, we simply replace $G$ with $\tilde G$ in the definition
of $C_h$, and use a diagonal reduction. This leads to $C_h\approx
\Diag(\tilde G J_\theta^{-1}\transp{\tilde G\,})$ and
\begin{equation}
J_h \approx \left(\Diag(\tilde G J_\theta^{-1}\transp{\tilde G\,})\right)^{-1}
\end{equation}
where as usual $\tilde G$ is the gradient approximation given by
\eqref{eq:Gtilde}.

The diagonal reduction is necessary if $\tilde G$ is low-rank, since
$\tilde G J_\theta^{-1} \transp{\tilde G\,}$ will be low-rank as well,
and thus non-invertible.

Then the scaling factors $\bar \rho$ and $\rho_i$ can finally be computed as
\begin{equation}
\bar\rho=\sqrt{\frac{\norm{\bar w}}{\norm{\bar v}}}
=\frac{(\transp{\bar w} J_\theta^{-1} \bar w)^{1/4}}{\left(\sum_i (\tilde G
J_\theta^{-1}\transp{\tilde G\,})^{-1}_{ii} \bar v_i^2\right)^{1/4}}
\end{equation}
and
\begin{equation}
\rho_i=\sqrt{\frac{\norm{w_i}}{\norm{e_i}}}
=\frac{(\transp{w_i} J_\theta^{-1} w_i)^{1/4}}{\left((\tilde
G
J_\theta^{-1}\transp{\tilde G\,})^{-1}_{ii} \right)^{1/4}}
\end{equation}
The particular structure of
$J_\theta$ (if approximated by, e.g., a block-diagonal matrix) and of $\tilde G=\bar
v\transp{\bar w}+\sum_i e_i \transp{w_i}$ make
these computations efficient.

Note that even with the approximations above, $\tilde G$ is still an
unbiased estimate of $G$. Indeed, any choice of $\rho$ has this property;
we are simply approximating the optimal $\rho$ which minimizes the variance of
$\tilde G$.

In practice, small regularization terms are included in the denominator
of every division and inversion to avoid
numerical overflow.
}

}%appendix

\bibliographystyle{alpha}
\bibliography{nobacktrack}

\begin{thebibliography}{BEYY04}

\bibitem[BEYY04]{elyaniv_markov2004}
Ron Begleiter, Ran El-Yaniv, and Golan Yona.
\newblock On prediction using variable order markov models.
\newblock {\em Journal of Artificial Intelligence Research}, pages 385--421,
  2004.

\bibitem[Cap11]{Cappe_onlineHMM}
Olivier Capp{\'e}.
\newblock Online {EM} algorithm for hidden {M}arkov models.
\newblock {\em J. Comput. Graph. Statist.}, 20(3):728--749, 2011.

\bibitem[Hay04]{Haykin_book}
Simon Haykin.
\newblock {\em Kalman filtering and neural networks}.
\newblock John Wiley \& Sons, 2004.

\bibitem[Jae02]{Jaeger_tutorial}
Herbert Jaeger.
\newblock Tutorial on training recurrent neural networks, covering {BPTT},
  {RTRL}, {EKF} and the {\textquoteleft}{\textquoteleft}echo state
  network{\textquoteright}{\textquoteright} approach.
\newblock Technical Report 159, German National Research Center for Information
  Technology, 2002.

\bibitem[Oll15a]{gradnn}
Yann Ollivier.
\newblock Riemannian metrics for neural networks {I}: feedforward networks.
\newblock {\em Information and Inference}, 4(2):108--153, 2015.

\bibitem[Oll15b]{pcnn}
Yann Ollivier.
\newblock Riemannian metrics for neural networks {II}: recurrent networks and
  learning symbolic data sequences.
\newblock {\em Information and Inference}, 4(2):154--193, 2015.

\end{thebibliography}

\end{document}